\documentclass[letterpaper, 10 pt, conference]{ieeeconf} 
\usepackage{subfigure}
\usepackage{setspace}
\usepackage{graphics, caption}
\usepackage{amssymb, amsmath, times}
\usepackage{bm}
\usepackage{multirow}
\usepackage{pifont}
\usepackage[usenames,dvipsnames,svgnames,table]{xcolor}
\usepackage[export]{adjustbox}
\usepackage{footnote}
\usepackage{soul}
\usepackage{epstopdf, tikz}
\usetikzlibrary{shapes,arrows, circuits, patterns, snakes, decorations.pathreplacing}
\DeclareGraphicsExtensions{.eps}
\usepackage{schemabloc, verbatim, neuralnetwork}
\usepackage{cite}
\usepackage{color}
\colorlet{linkequation}{blue}
\colorlet{linkfigure}{green}
\usepackage[colorlinks=true, citecolor=green,linkcolor=blue,urlcolor=blue]{hyperref}
\usepackage{breqn}
\usepackage{algorithmicx}
\usepackage[capitalise]{cleveref}
\pdfoutput=1    
\tikzstyle{block} = [draw, fill=white!20, rectangle, 
    minimum height=3em, minimum width=4em]
\tikzstyle{sum} = [draw, fill=white!20, circle, node distance=1cm]
\tikzstyle{input} = [coordinate]
\tikzstyle{output} = [coordinate]
\tikzstyle{pinstyle} = [pin edge={to-,thin,black}]

\newcounter{mnote}

\newcommand{\ie}{i.e.\ }
\newcommand{\eg}{e.g.\ }

\newcommand{\flabel}[1]{\label{fig:#1}}

\newcommand{\elabel}[1]{\label{eq:#1}}

\newcommand*\idx[2][]
{ 
\def\next{#1}%
\ifx\empty\next
  (#2)
\else
  (#1, #2)
\fi
}
\newcommand*\elt[3][]
{ 
\def\next{#1}%
\ifx\empty\next
  #2\idx{#3}
\else
  #1\idx{#2,#3}
\fi
}
\newcommand*\pd[3][]
{ 
\def\next{#1}%
\ifx\empty\next
  \frac{\partial#2}{\partial #3}
\else
  \frac{{\partial^{#1} #2}}{\partial#3^{#1}}
\fi
}

\newcommand{\hiddenfn}{\mathcal{H}}

\newcommand{\wtmat}[2]{W_{#1 #2}}
\newcommand{\ihwts}{\wtmat{u}{h}}
\newcommand{\hhwts}{\wtmat{h}{h}}
\newcommand{\howts}{\wtmat{h}{y}}
\newcommand{\bias}[1]{b_{#1}}
\newcommand{\hbias}{\bias{h}}
\newcommand{\obias}{\bias{y}}
\newcommand{\seq}[1]{\boldsymbol #1}

\newcommand{\outvble}{y}

\newcommand{\outseq}{\seq{\outvble}}

\newcommand{\capt}[2]{\caption[#1]{#1#2}}

\newcommand{\figt}[5]
{
\begin{figure}[t]
\begin{center}
\includegraphics[width=#3\columnwidth]{figures/#1}
\end{center}
\capt{#4}{#5}
\flabel{#2}
\end{figure}
}

\title{\LARGE \textbf Nonlinear Systems Identification Using Deep Dynamic Neural Networks} 

\author{Olalekan Ogunmolu$^{1}$, Xuejun Gu$^{2}$, Steve Jiang$^{2}$, and Nicholas Gans$^{1}$   
	\thanks{*This work was supported by the Radiation Oncology Department, UT Southwestern, Dallas, Texas, USA}
	\thanks{$^{1}$Olalekan Ogunmolu and Nicholas Gans are with the Department of Electrical Engineering,
		University of Texas at Dallas, Richardson, TX 75080, USA
		{\tt\small \{olalekan.ogunmolu, ngans\}@utdallas.edu}}%
	\thanks{$^{2}$Xuejun Gu and Steve Jiang are with the Department of Radiation Oncology,  
		University of Texas Southwestern Medical Center, Dallas TX 75390, USA
		{\tt\small \{Xuejun.Gu, Steve.Jiang\}@utsouthwestern.edu}}%
}

\IEEEoverridecommandlockouts

\begin{document}
	
	
	\maketitle
	\thispagestyle{empty}
	\pagestyle{empty}

\begin{abstract}
Neural networks are known to be effective function approximators. Recently, deep neural networks have proven to be very effective in pattern recognition, classification tasks and human-level control to model highly nonlinear real-world systems. This paper investigates the effectiveness of deep neural networks in the modeling of dynamical systems with complex behavior. Three deep neural network structures are trained on sequential data, and we investigate the effectiveness of these networks in modeling associated characteristics of the underlying dynamical systems. We carry out similar evaluations on select publicly available system identification datasets. We demonstrate that deep neural networks are effective model estimators from input-output data.
\end{abstract}

\section{Introduction}

Methods for the adaptive identification and control of linear, time invariant systems with unknown parameters are well-established and documented in linear systems theory, with stable adaptive laws for the  adjustment of parameters that demonstrate global stability of the overall system. Being universal approximators, neural networks (NNs) have witnessed a flurry of use in modeling various nonlinear phenomena in the past three decades. Three broad classes of NNs that have received attention recently include 1) multilayer perceptrons, 2) recurrent neural networks, and 3) convolutional neural networks. Multilayer networks have been used in identification and control of static and dynamic simple nonlinear systems \cite{NarendraNNDynSys, NarendraIdentControl} while recurrent networks (and its variants) have been used as associative memories for the solution of time-series/sequential optimization problems \cite{Sutskever2013, Graves2013} and in the dynamic identification and control of nonlinear systems \cite{FARNN2006, Dinh2010}. Convolutional networks, on the other hand, have been successfully used in pattern recognition, supervised classification tasks and image processing problems \cite{LeCun98, krizhevsky2012}.

In complicated real-world systems, deep neural networks (DNNs) have proven very effective for classification problems related with patterns in complicated systems such as image processing  \cite{krizhevsky2012}, speech processing \cite{Sejnowski1987}, language models \cite{Zaremba2014}, handwriting recognition \cite{LeCun98} and sequential data \cite{Sutskever2013, Hochreiter1997}. These networks are termed ``deep" because they are constructed by stacking multiple layers of non-linear operations (such as NNs) atop one another with many hidden layers. They are analogous to complicated formulae that re-use many sub-formulae in abstracting real-world representations with their parameters (or weights). 

The work discussed in this paper is largely motivated by the problem discussed in recent investigations of the identification and control of soft-robots for head and neck motion alignment during cancer radiotherapy (RT) \cite{Case2015, Case2016}. Here we design self-organizing networks, connected in a DNN fashion, to enable the development of efficient and synaptic adaptive rules for arbitrarily connected NNs; this facilitates the development of an internal structure that is appropriate for a system identification and control learning task. 

This work presents NN-based Hammerstein models evaluated on SISO and MIMO datasets. The modeling procedure for approximating systems such as the ones we present in this work is a complicated task with highly nonlinear dynamics that may be too complicated to model with closed-form equations. We extend the  development of NNs for abstracting complex nonlinear real-world systems in the pattern recognition field over the past 2 decades to solving a recursive identification, parameter estimation and control problem of a complex system.

Three specific NN architectures are investigated namely the multilayer network, simple recurrent NN and its long short-term memory (LSTM) variants, encoded in various suitable architectures appropriate to our learning task 
To demonstrate the applicability and extensibility of this identification methods, we conduct separate identification experiments to test the effectiveness of these modeling procedures on select SISO- and MIMO-system identification datasets from  DaISy \footnote{DaISy: Database for the Identification of Systems by De Moor B.L.R.",
Department of Electrical Engineering, ESAT/STADIUS, KU Leuven, Belgium.  "http://homes.esat.kuleuven.be/~smc/daisy/"}. 


\section{Preliminaries and basic concepts} \label{sec:prelim}
The underlying principle in artificial NN models are an adaptation of the natural network of neurons originally proposed by \cite{Rosenblatt1961principles}, whereby each single neuron predicts an output by \textit{weighing} up the evidence of ``truths" from fed inputs and shifting the gradient of the resulting function based on an additive `\textit{bias}' term; a \textit{squashing} unit applies a nonlinear transformation to the linearly combined inputs to produce a desired bounded, and constant nonlinear output, $\hat{y}(t)$. By combining a large sum of these simple component connections across the input space and forwarding them through the layers of the network neuron nodes, we obtain a function $\hat{f} \subset D_f$, which uniformly approximates the continuous function $f: D_f \subset \mathbb{R}^{n_u} \rightarrow \mathbb{R}^{n_y} $ to an acceptable bounded error, $|\epsilon|$, where $D_f$ is a compact subset of $\mathbb{R}^{n_u}$, given that there are enough nodes in the network layers. 

The input-output relation of the system can be described by the following equation
\begin{align} \label{eq.neuron}
z_j^l (k) = f(\sum\limits_{i = 1}^{n}w_{ij}^l {x_i}^{l-1}(k) + \delta_i^l)
\end{align}
where $f(\cdot)$ is the nonlinear activation function, $w_{ij}^l$ is the connection weight of $j$th neuron units in the $(l-1)$th layer to those of the $l$th layer, $x_i^{l-1}$ is the input from the $(l-1)$th layer, $\delta_i^l$ are the respective reconstruction errors, or \textit{biases}, and $z_j^l$ is the output of the $j$th neuron in the $l$th layer for $i = 1, \ldots , n$. Common nonlinear activation functions used in practice include the logistic sigmoid function, 
$
	\sigma(x) = \dfrac{1}{1+e^{-x}},
$
the hyperbolic tangent function, 
$
	tanh(x) = \dfrac{e^{2x}-1}{e^{2x}+1},
$
or the point-wise rectified linear units, $max(0, x)$, where $x$ is the input. The logistic function is relevant to functions that map into probability output spaces while the hyperbolic tangent function maps to the output range $[-1, 1]$; the ReLU function has the advantage of being easier to optimize, providing faster convergence in networks, being easier to generalize and having a lower computational overhead \cite{Zeiler2013}. \footnote{Other versions of the ReLU exist such as the ReLU6 which is useful for training networks that do not loose precision; the parametric ReLU, where the negative gradients are learned from data rather than from being initialized at training time; and the leaky ReLU among others.} 

The goal in the identification of a nonlinear process is to find a mathematical representation given input-output data alone. Typically, a model of the system to be identified is expressed as an operator $F$ from an input space $\mathbb{U}$ to an output space $\mathbb{Y}$, and the goal is to find a function $\hat{F}$ that approximates $F$ to a specific requirement. A static NN would generally map from an input $\mathbb{U} \in \mathbb{R}^n $ to an output $\mathbb{Y} \in \mathbb{R}^m$, while a dynamic NN will map from an input $\mathbb{U}$ in a compact space to an output $\mathbb{Y}$ that is assumed to be bounded on the Lebesgue integrable functions on the closed interval $[0, T]$ or open-ended interval $[0, \infty)$. By the Stone-Weierstrass theorem, there exists a  continuous function, $F$, on the bounded, compact input space with interval $[a,b]$ $\subset \mathbb{{U}}$ such that for any $\epsilon > 0$ there is a function $f \in F$ that for all $u \in \mathbb{U}$, makes $|Fu - fu| < \epsilon$.

\section{Related Work}\label{sec:related-work}
Narendra \cite{NarendraIdentControl} proposed and justified models for the identification and control of low-order, bounded output nonlinear systems using static and dynamic feed-forward and recurrent NNs. Parameters of the network were adjusted through dynamic back-propagation. A drawback was that such networks were assumed stable, and the models they generated were assumed to be controllable, observable and identifiable.  In \cite{NarendraNNDynSys}, he showed that NNs were effective for the identification and control of multi-variable, higher-order complex dynamical systems. Wang et. al.\cite{FARNN2006} combined a static feed-forward network and a dynamic recurrent NN in an ad-hoc Hammerstein block-structured model to construct a greedy network that provided the automated identification of the underlying unknown model; this involved a careful initialization of the network weights and biases as proposed by \cite{Yam2001}, and they reported achieving convergence faster (compared against random initialization of parameters) during training by avoiding the problem of saturated sigmoid activation layers.

An appropriate cost function to be optimized maps the low-dimensional features to the output space. The time and effort given to careful weights initialization (such as simulated annealing or genetic algorithms) make training difficult and not easily generalizable to new datasets, as best initial weights have to be carefully selected for every new problem. Even so, in recurrent NNs, the temporal dependence of network parameters that are computed based on  previous weight matrices raised to a high power cause gradients to grow or vanish proportionally to the exponent of the number of previous temporal steps \cite{Bengio1994}.  The saturation of neurons in the hidden layer also increases training time by a significant factor. Therefore, hand-coding features is an elaborate task that makes recurrent NN training difficult, as most literature show.

Self-organizing NNs, proposed by LeCun \cite{LeCun98}, with deep architectures find adaptive and automatic learning rules that enable connected NNs to develop an internal structure that is appropriate for a particular nonlinear parameter estimation (\textit{learning}) task. LeCun showed that the traditional methods of designing hand-coded features for recognition systems can be replaced by training component-wise modules that collectively work together to optimize a global performance criterion.


\section{Supervised Learning with Neural Networks}
To allow an arbitrarily connected NN to develop self-adaptive learning rules that model an unknown system based on a finite data set, $Z^N$ (consisting of input-output pair, $\{u_1, u_2, \cdots u_N, y_1 \cdots , y_N\}$), we use a 
network topology that learns rules for adjusting the network weights $W_i$ in order to make the predicted outputs $\hat{y}_i(k)$ approximate the desired outputs $y_i(k)$ to a sufficient degree, $\epsilon$. Cybenko \cite{Cybenko89} and Funahashi \cite{Funahashi89} have shown that a single hidden layer is sufficient as a universal function approximator with the ability to approximate any Borel measurable function from one finite dimensional space to another. A single hidden layer can achieve a sufficient degree of accuracy with no theoretical constraint on the network's learning ability 
\cite{Hornik1989}. The presence of noise in data can make a NN optimization get stuck in local minima during backpropagation but deep networks are better at identifying model structure in data in spite of noise in data. 

In a typical feedforward NN, the input data is fed into an input layer that distributes the data to hidden layer(s), consisting of neurons that connect to the neurons of other layers; the NN may contain more than one hidden layer, but the signals from the last hidden layer must flow toward that of the output layer. The parameters of the network are chosen to minimize a global loss function $Q(z, \hat{{y}}) = l(\hat{y}, y)$, which measures the cost of predicting the $\hat{y}$ when the true output ${y}$ is a function over the training set. For regression problems encountered in system identification tasks, it is typical to use the mean-squared error as a cost to be minimized, $\ie$,
\begin{align}  \label{eq:MSE}
 l(\hat{y}, y) = \sum\limits_{k = 1}^{K}E(k) = \frac{1}{2n}\sum\limits_{k = 1}^{K}\sum\limits_{i = 1}^{n}||\hat{y}_i(k) - y_i(k)||^2
\end{align} 
using the basic backpropagation algorithm \cite{Rumelhart1986} for a feedforward network or the popular backpropagation through time for a recurrent NN \cite{Werbos1990}. \eqref{eq:MSE} is a special case of the least-squares method, with $n$ being the total number of training examples; $l(\hat{y}, y)$  is minimized over the training examples using gradient descent so that at each iteration, we update the parameters $w_i$ based on the gradient of $Q(z, \hat{y})$ $\ie$,
\begin{align} \label{eq.GDUpdate}
	w_{k+1} \leftarrow \eta w_k - \alpha \frac{1}{n}\sum\limits_{i=1}^{n}\nabla_w Q(z_i, \hat{y}_i(w_k)),
\end{align}
where $\eta$ is the momentum that speeds up the optimization along directions of low but persistent reduction in training error \cite{krizhevsky2012},  $\alpha$ is a sufficiently small learning rate, and $\nabla_w Q(z, w_k)$ is the derivative of $Q$ with respect to $w$. If $\nabla_w Q(z, w_k)$ $\neq0$, then for sufficiently small and positive definite $\alpha$,
$\eta \, w_k - \alpha \frac{1}{n}\sum\limits_{i=1}^{n}\nabla_w Q(z_i, w_k) < w_0$. Therefore, \eqref{eq.GDUpdate} has linear convergence under sufficient regularity assumptions when the starting point $w_0$ is close enough to the minimum value of the loss. In practice, a simplification of \eqref{eq.GDUpdate}, termed stochastic gradient descent (SGD), is used in computing an estimate of the gradient based on a single randomly picked example $z_k$
\begin{align} \label{eq.SGDupdate}
w_{k+1} \leftarrow \eta w_k - \alpha_k\nabla_w Q_k(z_k, w_k),
\end{align}
where $\nabla_w Q_k(\cdot)$ is the average over the $k$-th batch of $\nabla_w Q$. \eqref{eq.SGDupdate} randomly samples from the training set during each epoch and directly optimizes $l(\hat{y}, y)$. 
SGD has the advantage of minimizing training time by computing an approximation to the gradient over each mini-batch of samples.

\section{Learning with Deep Dynamic NNs} \label{sec:learning}
The datasets considered in this work are sequential in nature, some with temporal correlation in the evolution of inputs. We therefore propose NN architectures that are adept at learning the nonlinearity in time-series data. Specifically, we consider feedforward multilayer networks, simple recurrent networks, long short-term memory and gated recurrent units. 

\begin{figure} [tb!]
	\centering
	\begin{tikzpicture}[thick,scale=0.5, every node/.style={transform shape}]
		\sbEntree{u_top}
		\sbStyleBloc{text=black, circle}
		\sbBlocL[5]{u1}{}{u_top}
			\sbRelier[$\textbf{u}_1(k)$]{u_top}{u1}				
		\sbDecaleNoeudy[9]{u_top}{u_down}
		\sbDecaleNoeudy[9]{u1}{u_m}	
		\sbStyleBloc{text=black, circle}
		\sbBlocL[5]{u_n}{}{u_down}
			\sbRelier[$\textbf{u}_n(k)$]{u_down}{u_n}											
		\sbDecaleNoeudy[3]{u1}{u1_trunk}
		\sbDecaleNoeudy[-3]{u_n}{un_trunk}
		\draw[dotted, color=gray](u1_trunk) -- (un_trunk);
		
		\sbDecaleNoeudy[-3]{u1}{u1h}
		\sbStyleBloc{text=black, ellipse}
		\sbBloc[7]{h11}{$\Sigma$}{u1h}			
		\sbDecaleNoeudy[3]{u1}{u2h}		
		\sbBloc[7]{h12}{$\Sigma$}{u2h}			
		\sbDecaleNoeudy[3]{u_n}{unh}
		\sbBloc[7]{h1n}{$\Sigma$}{unh}
		
		\sbRelier[$\textbf{w}_{n1}$]{u_n}{h11}
		\sbRelier[]{u_n}{h12}
		\sbRelier[$\textbf{w}_{nn}$]{u_n}{h1n}	
		\sbRelier[$\textbf{w}_{11}$]{u1}{h11}
		\sbRelier[]{u1}{h12}
		\sbRelier[]{u1}{h1n}
		
		\sbSortie[4]{x1}{h11}
		\sbNomLien[1.2]{x1}{$\textbf{x}_1$}
		\sbSortie[4]{x2}{h12}
		\sbNomLien[1.2]{x2}{$\textbf{x}_2$}
		\sbSortie[4]{xn}{h1n}
		\sbNomLien[1.2]{xn}{$\textbf{x}_n$}
		
		\sbDecaleNoeudy[3]{x2}{x2n}
		\sbDecaleNoeudy[-3]{xn}{xnn}
		\draw[dotted, color=gray](x2n) -- (xnn);
		
		\sbSortie[3]{fx1}{h11}
		\sbStyleBlocDefaut 
		\sbBlocL[5]{fx1}{$\textbf{f}(.)$}{h11}	
		\sbSortie[3]{fx2}{h12}
		\sbStyleBlocDefaut 
		\sbBlocL[5]{fx2}{$\textbf{f}(.)$}{h12}
		\sbSortie[3]{fxn}{h1n}
		\sbStyleBlocDefaut 
		\sbBlocL[5]{fxn}{$\textbf{f}(.)$}{h1n}
						
		\sbDecaleNoeudy[3]{fx1}{x12}
		\sbStyleBloc{circle}
		\sbBloc[8]{x1x2}{$\Sigma$}{x12}
		\sbDecaleNoeudy[3]{fx2}{x2n}
		\sbBloc[8]{x2xn}{$\Sigma$}{x2n}
		
		\sbRelier[$\bar{\textbf{w}}_{11}$]{fx1}{x1x2}	
		\sbRelier[]{fx1}{x2xn}		
		\sbRelier[]{fx2}{x1x2}
		\sbRelier[]{fx2}{x2xn}
		\sbRelier[$\bar{\textbf{w}}_{n1}$]{fxn}{x1x2}
		\sbRelier[]{fxn}{x2xn}
		
		\sbSortie[4]{y1}{x1x2}
		\sbRelier{x1x2}{y1}
		\sbNomLien[1.2]{y1}{$\textbf{y}_1(k)$}
		\sbSortie[4]{ym}{x2xn}
		\sbRelier{x2xn}{ym}
		\sbNomLien[1.2]{ym}{$\textbf{y}_m(k)$}
		
		\sbDecaleNoeudy[1]{y1}{y1n}
		\sbDecaleNoeudy[-2]{ym}{ynn}
		\draw[dotted, color=gray](y1n) -- (ynn);
	\end{tikzpicture}
	\caption{MLP network.} 
	\label{fig:mlp-sr}
\end{figure}
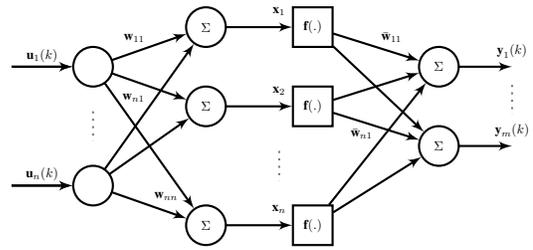

\subsection{Multilayer Networks}
\autoref{fig:mlp-sr} shows the schematic representation of a multilayer NN (MLP) 
with synchronous signals that flow in a forward direction, $U \rightarrow H \rightarrow Y$. Joining the weights and biases of the network completely parameterizes the system it is trained on.
During training, the estimated outputs are compared with the true outputs to calculate the error signal in the network. The errors are back-propagated through the network to obtain the \textit{ordered} derivatives for learning. The \textit{``goodness"} of the trained model can be measured by evaluating how well the training data generalizes to testing data which is separated from the training set. 

\subsection{Recurrent Neural Networks} \label{sec:recurrent}
\begin{figure}[tb!] 
	\centering
	\begin{tikzpicture}[thick,scale=0.5, every node/.style={transform shape}]
		\sbEntree{u_top}
		\sbStyleBloc{text=black, circle}
		\sbBlocL[5]{u1}{}{u_top}
			\sbRelier[$\textbf{u}_1(k)$]{u_top}{u1}				
		\sbDecaleNoeudy[9]{u_top}{u_down}
		\sbDecaleNoeudy[9]{u1}{u_m}	
		\sbStyleBloc{text=black, circle}
		\sbBlocL[5]{u_n}{}{u_down}
			\sbRelier[$\textbf{u}_n(k)$]{u_down}{u_n}											
		\sbDecaleNoeudy[3]{u1}{u1_trunk}
		\sbDecaleNoeudy[-3]{u_n}{un_trunk}
		\draw[dotted, color=gray](u1_trunk) -- (un_trunk);
		
		\sbDecaleNoeudy[-3]{u1}{u1h}
		\sbStyleBloc{text=black, ellipse}
		\sbBloc[7]{h11}{$\Sigma$}{u1h}			
		\sbDecaleNoeudy[3]{u1}{u2h}		
		\sbBloc[7]{h12}{$\Sigma$}{u2h}			
		\sbDecaleNoeudy[3]{u_n}{unh}
		\sbBloc[7]{h1n}{$\Sigma$}{unh}
		
		\sbDecaleNoeudy[3]{u_down}{ustart}
		\sbDecaleNoeudy[3]{u_n}{uend}
		\draw[thick, decoration={brace, mirror, raise=0.5cm},
		decorate](ustart) -- (uend)
		node[pos=0.5, anchor=north, yshift=-1.2cm]{\textbf{\Large input layer}};
		
		\sbRelier[]{u_n}{h11}
		\sbRelier[]{u_n}{h12}
		\sbRelier[]{u_n}{h1n}	
		\sbRelier[]{u1}{h11}
		\sbRelier[]{u1}{h12}
		\sbRelier[]{u1}{h1n}
		
		\sbSortie[4]{x1}{h11}
		\sbNomLien[1.2]{x1}{$\textbf{x}_1$}
		\sbSortie[4]{x2}{h12}
		\sbNomLien[1.2]{x2}{$\textbf{x}_2$}
		\sbSortie[4]{xn}{h1n}
		\sbNomLien[1.2]{xn}{$\textbf{x}_n$}
		
		\sbDecaleNoeudy[3]{x2}{x2n}
		\sbDecaleNoeudy[-3]{xn}{xnn}
		\draw[dotted, color=gray](x2n) -- (xnn);
			
		\sbSortie[3]{fx1}{h11}
		\sbStyleBlocDefaut 
		\sbBlocL[5]{fx1}{$\textbf{f}(.)$}{h11}	
		\sbSortie[3]{fx2}{h12}
		\sbStyleBlocDefaut 
		\sbBlocL[5]{fx2}{$\textbf{f}(.)$}{h12}
		\sbSortie[3]{fxn}{h1n}
		\sbStyleBlocDefaut 
		\sbBlocL[5]{fxn}{$\textbf{f}(.)$}{h1n}
		
		\sbDecaleNoeudy[-1]{h11}{fdpt1}
		\sbDecaleNoeudy[-1]{h12}{fdpt2}
		\sbDecaleNoeudy[-1]{h1n}{fdpt3}
		\draw [->] (fx1) to [out=120,in=60] (fdpt1);
		\draw [->] (fx2) to [out=120,in=60] (fdpt2);
		\draw [->] (fxn) to [out=120,in=60] (fdpt3);
		\draw [->] (h11) to [out=150,in=210] (h12);
		\draw [->] (h11) to [out=150,in=210] (h1n);
		\draw [->] (h12) to [out=150,in=210] (h1n);
		\draw [->] (h12) to [out=150,in=210] (h11);
		\draw [->] (h1n) to [out=150,in=210] (h11);
		\draw [->] (h1n) to [out=150,in=210] (h12);
		
		\sbDecaleNoeudy[-0.8]{x1}{sstart}
		\sbSortie[0.34]{fx11}{fx1}
		\sbDecaleNoeudy[-0.8]{fx11}{send}
		\draw[thick, decoration={brace, raise=0.32cm},
		decorate](sstart) -- (send)
		node[pos=0.7, anchor=north, yshift=1.5cm]{\textit{\textbf{\Large squashing units}}};

		\sbDecaleNoeudy[0.1]{unh}{hstart}
		\sbDecaleNoeudy[.1]{fxn}{hend}
		\draw[thick, decoration={brace, mirror, raise=0.5cm},
		decorate](hstart) -- (hend)
		node[pos=0.5, anchor=north, yshift=-1.2cm]{\textbf{\Large hidden layer}};
								
		\sbDecaleNoeudy[3]{fx1}{x12}
		\sbStyleBloc{circle}
		\sbBloc[8]{x1x2}{$\Sigma$}{x12}
		t
		the input gate allows the memory unit to be updated
		mt = mt ?ot
		˜
		the output gate determines if information can leave the unit
		zt = g(Wyhht +Wym ˜ mt)
		(2.37)x2DaISy and xn
		\sbDecaleNoeudy[3]{fx2}{x2n}
		\sbBloc[8]{x2xn}{$\Sigma$}{x2n}
		
		\sbRelier[$\bar{\textbf{w}}_{11}$]{fx1}{x1x2}	
		\sbRelier[]{fx1}{x2xn}		
		\sbRelier[]{fx2}{x1x2}
		\sbRelier[]{fx2}{x2xn}
		\sbRelier[$\bar{\textbf{w}}_{n1}$]{fxn}{x1x2}
		\sbRelier[]{fxn}{x2xn}
		
		\sbSortie[4]{y1}{x1x2}
		\sbRelier{x1x2}{y1}
		\sbNomLien[1.2]{y1}{$\textbf{y}_1(k)$}
		\sbSortie[4]{ym}{x2xn}
		\sbRelier{x2xn}{ym}
		\sbNomLien[1.2]{ym}{$\textbf{y}_m(k)$}
		
		\sbDecaleNoeudy[1]{y1}{y1n}
		\sbDecaleNoeudy[-2]{ym}{ynn}
		\draw[dotted, color=gray](y1n) -- (ynn);

		\sbDecaleNoeudy[6.2]{x2xn}{xstart}
		\sbDecaleNoeudy[8.2]{ynn}{xend}
		\draw[thick, decoration={brace, mirror, raise=0.5cm},
		decorate](xstart) -- (xend)
		node[pos=0.5, anchor=north, yshift=-1.2cm]{\Large \textbf{output layer}};		
	\end{tikzpicture}
	\caption{A simple recurrent neural network.} 
	\label{fig:rnn-simple}
	\vspace{-0.4cm}
\end{figure}
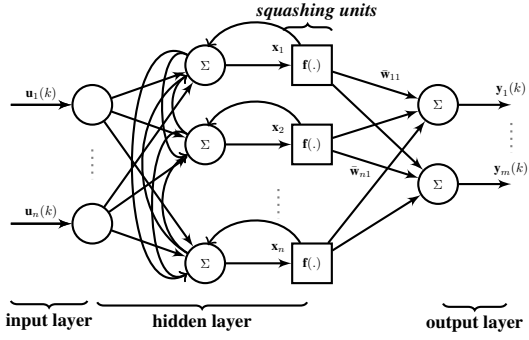

Recurrent Neural Networks (RNNs) are modeled from the behavior of many cells in nature with content-addressable memory, capable of capturing an entire information sequence given portions of the overall sequence. Whereas, the forward networks ``fire" their neurons in a single direction, RNNs employ a strong back-coupling $U { \rightleftarrows} H {\leftrightarrows} Y  { \rightleftarrows} U$ such that signal strengths can flow asynchronously between nodes even when a node signal is delayed. The architecture of a simple RNN is similar to that of a MLP, except that there is a self-feedback of neurons in the hidden layer(s) (see \autoref{fig:rnn-simple}). RNNs model nonlinear dynamical systems whose phase space dynamics is determined by a significant number of locally stable nodes to which it is attracted  \cite{Hopfield1982}. The hidden nodes $\seq{h} = (h_1,\ldots,h_N)$ and output nodes $\outseq = (y_1,\ldots,y_N)$ are determined by looping through the equations
\begin{align}
\elabel{rnn_hidden}
h_k &= \hiddenfn\left (\ihwts u_k + \hhwts h_{k-1} + \hbias \right)\\
y_k &= \howts h_k + \obias
\end{align} 
from $k=1$ to $N$ where  the $W$ terms are the weight matrices (\eg $\ihwts$ would be the input-to-hidden weight matrix), the $b$ terms represent the vectorized bias terms (\eg $\hbias$ would be the hidden bias vector) and $\hiddenfn$ is the hidden layer function, applied as an Hadamard operator. The loss is a cummulative loss of each time-step losses and the gradients are computed through backpropagation through time (BPTT \cite{Werbos1990, Rumelhart1986}) whereby parameters are updated after a complete sequence of forward and backward passes are completed or real-time recurrent learning (RTRL).

\subsection{Long Short Term Memory (LSTM) Cells} \label{sec:lstm}
For long-term context memorization, the gradients of RNNs can become intractable, as they use their back-coupling connections to \textit{memorize} the structure of recent inputs ($\ie$ short-term memory as compared against long-term memory). As a result,  backpropagated error signals in time can become infinitely high (causing oscillating weights), or vanish (causing complexity in computing slow varying weights) to the extent that the evolution in time of the backpropagated errors exponentially depend on the size of the weights \cite{Bengio1994, Hochreiter1997}.  
\begin{figure}[tb!] 
\centering
\includegraphics[width=3.2in, clip=true, center]{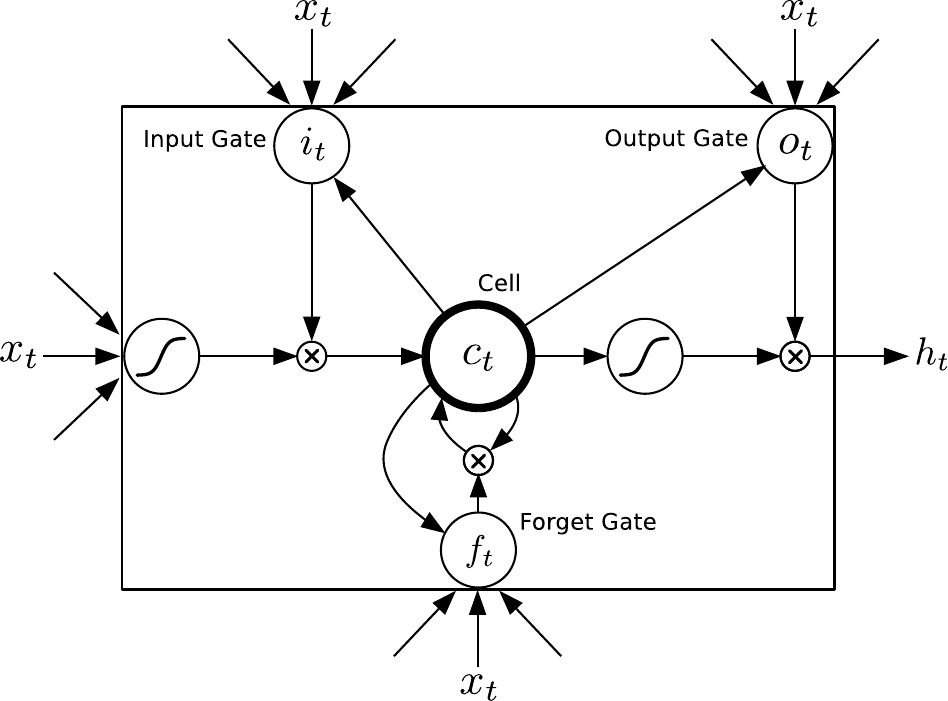}
\caption{Long Short-term Memory Cell. Reprinted from \cite{Graves2013}.}
\label{fig:lstm}
\vspace{-0.68cm}
\end{figure}
Horchreiter et al.(1997) \cite{Hochreiter1997} proposed the LSTM remedy that truncates gradients in the network where it is innocuous by enforcing constant error flows through \textit{constant error carousels} within special multiplicative units (MUs). Constant error flow is regulated by nonlinear MUs that learn to open or close gates in the network. LSTMs therefore approximate long-term information with significant delays by solving RNN algorithms faster. For an LSTM cell with $N$ memory units, at each time step, the evolution of its parameters are determined by 
\begin{align}
i_t &= \sigma(W_{u_i}u_t + W_{h_i}h_{t-1} + W_{c_i}c_{t-1}+b_{i_i}) \nonumber \\
f_t &= \sigma(W_{u_f}x_t + W_{h_f}h_{t-1} + W_{c_f}c_{t-1}+b_{i_f})  \nonumber \\
z_t &= tanh(W_{u_c}u_t + W_{h_c}h_{t-1} + b_{c})						\nonumber \\
c_t &= f_t \odot c_{t-1} + i_t \odot \, z_t								\nonumber 		\\
o_t &= \sigma(W_{u_o}u_t + W_{h_o}h_{t-1} + W_{c_o}c_{t-1}+b_{i_o}) \nonumber  \\
h_t &= o_t \odot tanh(c_t)
\end{align}
where the $W_{u_q}$ and $W_{h_q}$ terms are the respective rectangular input and square recurrent weight matrices, $W_{c_q}$ are peephole weight vectors from the cell to each of the gates (see \autoref{fig:lstm}), $\sigma$ denotes sigmoid activation functions (applied element-wise) and the $i_t$, $f_t$ and $o_t$ equations denote the input, forget and output gates respectively; $z_t$ is the input to the cell $c_t$. The output of the LSTM cell is $o_t$ and $\odot$ denote point-wise vector products. The bias terms for the gates are initialized to a large value at the beginning of training in order to allow learning long-term context \cite{Sutskever2013}. 
The forget gate facilitates resetting the state of the LSTM, while the peephole connections from the cell to the gates enable accurate learning of timings. 
%
\subsection{Fast LSTM} \label{sec:fastlstm}
This is a faster version of the LSTM architecture of \autoref{fig:lstm}, with the input, forget and the output gates of the LSTM cell computed without using the connections from the peepholes. The fast LSTM algorithm is computed as follows
\begin{align}
i_t &= \sigma(W_{x_i}x_t + W_{h_i}h_{t-1} + b_{i_i}) \qquad &\textit{input gate} \nonumber \\
f_t &= \sigma(W_{x_f}x_t + W_{h_f}h_{t-1} + b_{i_f}) \qquad &\textit{forget gate} \nonumber  \\
z_t &= tanh(W_{x_c}x_t + b_{c})	\qquad &\textit{block input}			\nonumber 		\\
c_t &= f_t \odot c_{t-1} + i_t \odot \, z_t	\qquad			&\textit{cell state}		\nonumber 				\\
o_t &= \sigma(W_{x_o}x_t + W_{h_o}h_{t-1} + b_{i_o}) \qquad &\textit{output gate} \nonumber  \\
h_t &= o_t \odot tanh(c_t) \qquad &\textit{block output}
\end{align}

\subsection{Gated Recurrent Units (GRU)} \label{sec:gru}
GRUs are simpler versions of LSTMs albeit with simpler computation of hidden states. They consist of two RNN systems acting in an encoder-decoder fashion: one RNN encodes a source sequence into a fixed-length vector representation, and the other RNN transforms the representations into a variable-length sequence whilst being jointly trained to maximize the conditional probability of a target sequence given an input sequence \cite{ChoGRU2014}. The GRU has a hidden state, $\textbf{h}_t$, that encodes the input sequence as a summary, $\textbf{c}$, while the decoder predicts the output sequence conditioned on previous outputs, $y_{t-i}$, and $\textbf{c}$ $\ie$
\begin{align}
&\textbf{h}_t = f(\textbf{h}_{t-1}, \textbf{y}_{t-1}, \textbf{c}) \nonumber \\
& \textbf{P}(y_t | y_{t-1}, y_{t-2}, \ldots , \textbf{c}) = g (\textbf{h}_t, y_{t-1}, \textbf{c})
\end{align}
where $f(\cdot)$ and $g(\cdot)$ are appropriate activation functions. Similar to the LSTM, the GRU has a hidden state that can forget previous information based on the state of a \textit{reset gate} as the following equations show
\begin{align}
r_i &= f \left([\textbf{W}_r\textbf{u}]_i + [\textbf{U}_r \textbf{h}_{t-1}]_i
\right) \qquad &\textit{reset gate} \nonumber \\
z_i &= f\left([\textbf{W}_z\textbf{u}]_i + [\textbf{U}_z \textbf{h}_{t-1}]_i\right) \qquad &\textit{update gate} \nonumber \\
h_i(t) &= z_i h_i(t-1) + (1-z_i)\bar{h}_i(t) \qquad &\textit{hidden activations} \nonumber \\
\bar{h}_i(t) &= g([\textbf{Wu}]_i + [\textbf{U}(\textbf{r}\odot \textbf{h}_{t-1})_i]
) 
\end{align}
where $f(\cdot)$ is a sigmoid activation function while $g(\cdot)$ can be activation functions that maps to probability spaces ($\eg$ soft-max).

\section{Experiments}
We develop and train models on the soft-robot dataset in \cite{Case2015, Case2016} and extend the results by training on select DaISy dataset.

\subsection{Data and Baseline Systems}
For the soft-robot actuator dataset, we had a mannequin head lying in a supine position on a table that simulated our proposed motion alignment correction set-up during cancer RT. A soft-robot actuator in the form of an inflatable air bladder (IAB) moved the mannequin head based on supplied air pressure. This corrected for non-rigid motions during treatment\cite{Case2015}. The IAB was actuated by current-driven proportional pneumatic valves; the experimental set-up is described in \cite{Case2015}, but the change in head motion is recorded by a motion capture (mocap) system instead of an RGB-D camera system. The mocap is capable of measuring head position with less than 1$mm$ error. This is a SISO system with input as current (generated from pseudo-random binary sequences) in $mA$ and outputs as head height in $mm$. We collected $10,070$ samples of input-output data offline, and in all experiments, we separate the dataset in a 60:40\% ratio for training and testing purposes. 
A mini-batch of 100 samples from the $\{u(k), y(k)\}$ data was used for a total of 50 epochs, where we loop over each mini-batch $10,000$ times and all training was performed on an NVIDIA CUDA-capable GPU.

In a separate experiments, we conducted training on the glassfurnace dataset which we downloaded from the DaISy file server. The glassfurnace dataset consists of 3 inputs and 6 outputs. The inputs are made up of two heating and a cooling signal, while the outputs are the readings from 6 temperature sensors in a cross-section of the furnace.

We examine the ability of deep dynamic NNs to model the underlying system dynamics using deep NN structures appropriate to the learning task guided by our knowledge of each system. We map a single input (being actuation current to the inlet pneumatic valve) to the pitch motion of the manikin head and allow the mass of the patient's head to naturally deflate the air bladder. For more complicated networks that we develop, such as recurrent and dynamic feedforward network Hammerstein models, we adopt \textit{dropout} techniques during training since the large number of parameters in the network could potentially lead to overfitting \cite{Hinton2014}. The code for replicating most of the experiments in this work can be found in \href{https://github.com/lakehanne/FARNN}{https://github.com/lakehanne/FARNN.}\footnote{Soft-Robot models are on the soft-robot branch; glassfurnace models are on the glassfurnace branch. An extensive discussion of the training procedure for other DaIsY datasets shall be posted on the author's blog at \href{http://lakehanne.github.io/}{http://lakehanne.github.io/}.}

\subsection{Multilayer network}
\subsubsection{Soft-Robot}
The current from the pneumatic valve was mapped to a hidden layer with six neurons, followed by a $ReLU$ nonlinearity that was then fully connected to the output layer ($\ie$ mocap measurements) (\autoref{fig:mlp-sr}). We conduct experiments with the ${current-pitch}$ data-pair (because the soft-robot directly controls the head pitch motion). 

The soft-robot multilayer network has 19 parameters; through cross-validation, we found a step size, $\alpha = \frac{1}{1000}$ to work well . We initially tried batch normalization of the hidden layer neurons and dropout regularization but these produced no noticeable speed-up in training time for the MLP network.  

\figt{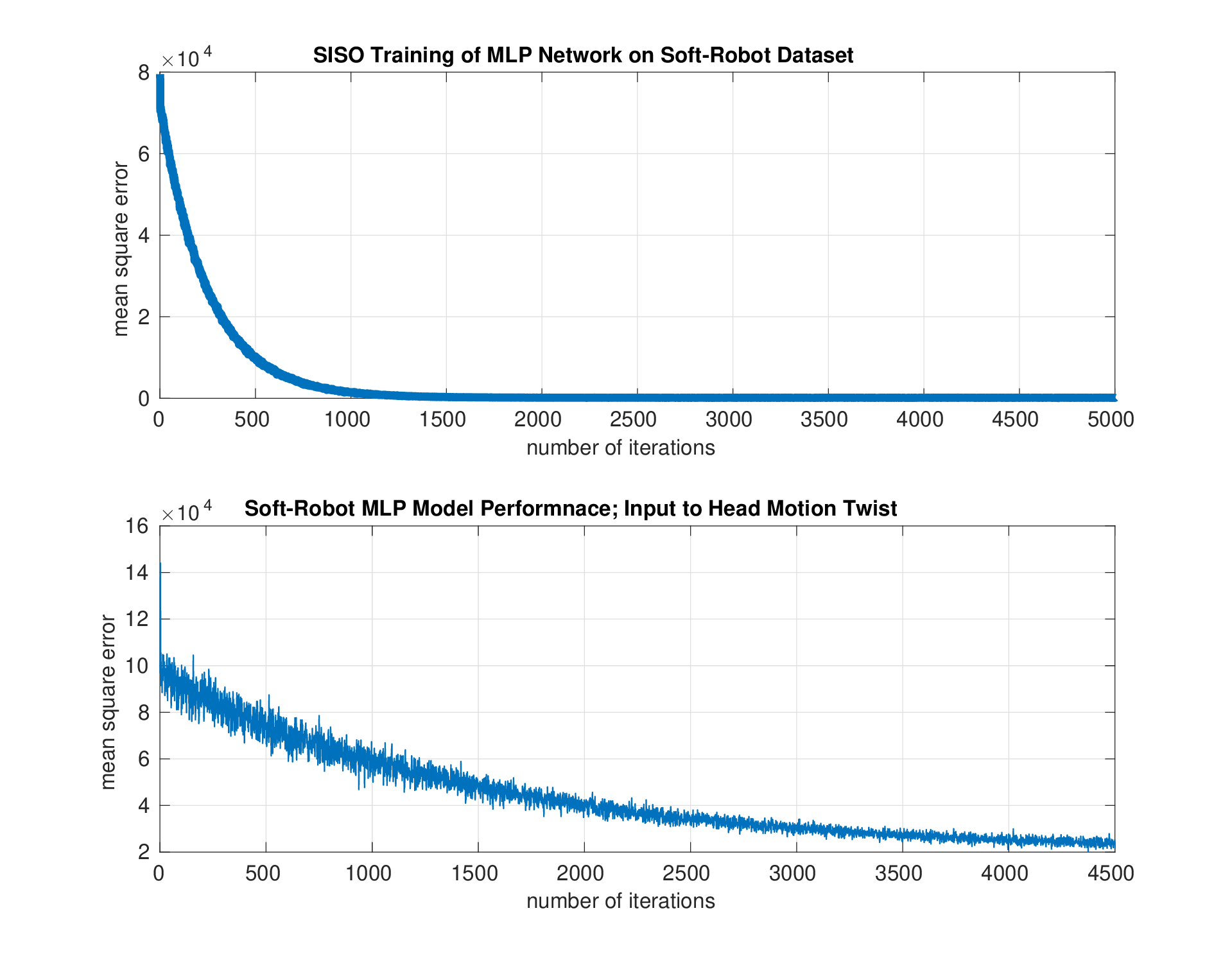}{mlp_mse}{}{\small Training of soft-robot system using a six-hidden layer MLP.}{ Fit to estimation data: SISO = 99.8\%, SIMO = 87.5 \%}
\subsubsection{GlassFurnace}
The model structure is similar to that of the SISO soft-robot system except that we use 3 input linear layers and we reshape the output layer to $6$. The performance is shown in the top chart of \autoref{fig:glass-rnn}.

\subsection{Recurrent Neural Network Structure}
\subsubsection{Soft-Robot}
\begin{figure}[tb]  
	\centering
	\begin{tikzpicture}[thick,scale=0.7, every node/.style={transform shape}]			
	\sbEntree{E}
	\sbBloc[4]{gdot}{$g(.)$}{E}
	\sbRelier[$u(n)$]{E}{gdot}
	\sbBloc{Hdot}{$G(z^{-1})$}{gdot}
	\sbRelier[$w$]{gdot}{Hdot}
	\sbCompSum[5]{sumpt}{Hdot}{}{}{}{}
	\sbRelier{Hdot}{sumpt}
	\sbSortie[3]{lastpt}{sumpt}
	\sbRelier[$y(n)$]{sumpt}{lastpt}
	\sbDecaleNoeudy[-4]{sumpt}{eta}
	\sbRelier[$\mu(n)$]{eta}{sumpt}
\end{tikzpicture}
	\caption{The Hammerstein model structure}
	\label{fig:hammerstein}
\end{figure}
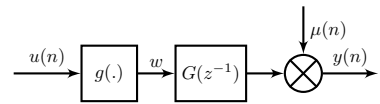

From our previous investigations of the soft-robot network, we had noticed a nonlinearity from input to states that the LTI models we earlier studied did not sufficiently capture \cite{Case2016}. We conjecture that a nonlinearity from input to system states followed by a dynamic linearity from states to head pitch motion would be favorable by feeding back interior nodes in the network as recurrent regressors. We propose a Hammerstein model consisting of a recurrent NN nonlinear element followed by a multilayer network to better model the overall system nonlinearity. The forward connections of the multilayer network would model the linear dynamic system from states to output (see \autoref{fig:hammerstein}). We employ this model structure with the three different recurrent network models discussed in section \autoref{sec:learning}, and  we map the valve current to head pitch motion. 
In the Hammerstein model of \autoref{fig:hammerstein}, the $g(\cdot)$ block represents the static nonlinearity that integrates the input sequence, and nonlinearly transforms the inputs to the system states. The neurons at this layer develop internal dynamics by their associative memory for $q$ steps back in time and weighted connections with the feedback connections from other neurons. The $G(z^{-1})$ block maps the linear dynamics of the system states to the sensors' measurements. We assume $g(\cdot)$ is continuous and bounded, and the linear dynamical system is causal and asymptotically stable. In all our models, we found a backpropagation in time by 5 steps ($\ie \, q = 5$ ) to be sufficient for approximating the system dynamics. We model the head motion of the patient as 
\begin{align}
y(n) = \dfrac{B(q^{-1})}{A(q^{-1})}g(u(n)) + \mu(n)
\end{align}
where $A(q^{-1})$ and $B(q^{-1})$ are regressive polynomials given by 
\begin{align}
A(q^{-1}) &= 1 + a_1 \, q^{-1} + \cdots + a_{n_a}q^{-n_a} \\
B(q^{-1}) &= b_1 \, q^{-1} + \cdots + b_{n_b}q^{-n_b}.
\end{align}
%
The $g(\cdot)$ and $G(z^{-1})$ networks are stacked on one another in a deep modular approach with weights updated along the negative gradients of the MSE cost function, $l(n)$. The parameters of the linear dynamic submodule, $\hat{a}_k(n)$, $\hat{b}_k(n)$, and the weight vector of the nonlinear element are updated according to 
\begin{align}
\hat{a}_k &= \hat{a}_k(n-1) - \eta \dfrac{\partial \, l(n)}{\partial \hat{a}_k(n-1)} \text{ , } k = 1, \ldots, n_a  \\
\hat{b}_k &= \hat{b}_k(n-1) - \eta \dfrac{\partial \, l(n)}{\partial \hat{b}_k(n-1)} \text{ , } k = 1, \ldots, n_b \\
w_c(n) &= w_c(n-1) - \eta \dfrac{\partial \, l(n)}{\partial w_c(n-1)} \text{ , } c = 1, \ldots, 3M + 1
\end{align}
with $\eta$ being the learning rate. The top graph of \autoref{fig:rnn-lstm} shows the performance of the RNN and feedforward multilayer Hammerstein network on the soft-robot dataset. This model performs faster and quickly integrates the mean-square error to reach the desired minimum compared to the forward network of the previous section. 
\subsubsection{Glassfurnace Data}
We adopt the same structure as the soft-robot network except that we reshape the input and output layers of the network model to accommodate the widths of the glassfurnace data. The training performance is depicted in \autoref{fig:glass-rnn}
\subsection{LSTM Model Architecture} \label{sec:lstm-training}
Exploiting the architecture of the recurrent network further, we replace the RNN nonlinear element of \autoref{fig:hammerstein} with the vanilla LSTM architecture discussed in \autoref{sec:lstm}. Our training model consists of three nonlinear LSTM modules, each decorated with dropout activation functions in their output layers, and the last layer being fully connected to a linear dynamic module. Note that this is a replication of the Hammerstein block-structured model. This is then fed to the vector of head motion measurements from the mocap system. 
\begin{figure}[tb!]
\centering
\includegraphics[width=3.6in, height=3.1in]{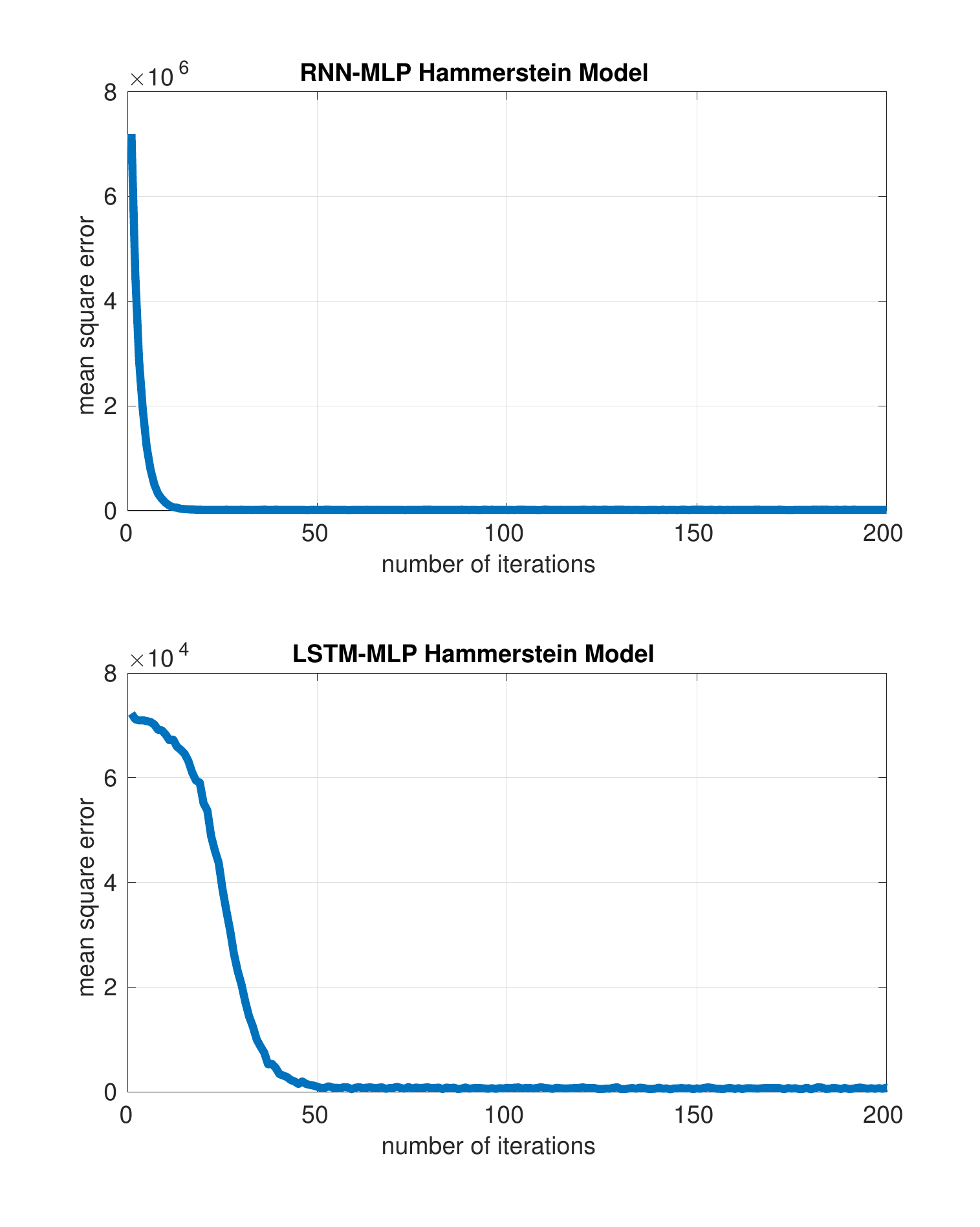}
\caption{\small RNN/Vanilla LSTM Hammerstein model performance on soft-robot dataset.}
\label{fig:rnn-lstm}
\end{figure}

Through model exploration, we found the following NN structure to work well with our dataset \eqref{eq:lstm-training}:
\begin{align}
    \text{Layer 1} &: \text{LSTM} & \quad 1 \rightarrow 1 \, \text{ \{0.3  dropout\}}	\nonumber \\
    \text{Layer 2} &: \text{LSTM} & \quad 1 \rightarrow 10 \, \text{ \{0.3  dropout\}}	\nonumber \\
    \text{Layer 3} &: \text{LSTM} & \quad 10 \rightarrow 100 \, \text{ \{0.3  dropout\}}	\nonumber \\
    \text{Layer 4} &: \text{Linear}& \quad  100 \rightarrow 1 . 	 
    \label{eq:lstm-training}
\end{align}
Altogether, the soft-robot network (SR) has $45,236$ while the glassfurnace dataset has $45989$ parameters. The SR performanace is shown in \autoref{fig:rnn-lstm}; this network handles input delay better given its capacity for modeling long-term dependencies as well as adapting its parameters to capture the temporal evolution of the underlying system. Training with the LSTM architecture takes a slightly longer time compared to the MLP or RNN-MLP architecture due to its highly recurrent nature and complexity in computing gradients. 

\subsection{Fast LSTM Architecture}
\begin{figure}[tbph!]
\centering
\includegraphics[width=3.6in, keepaspectratio=true]{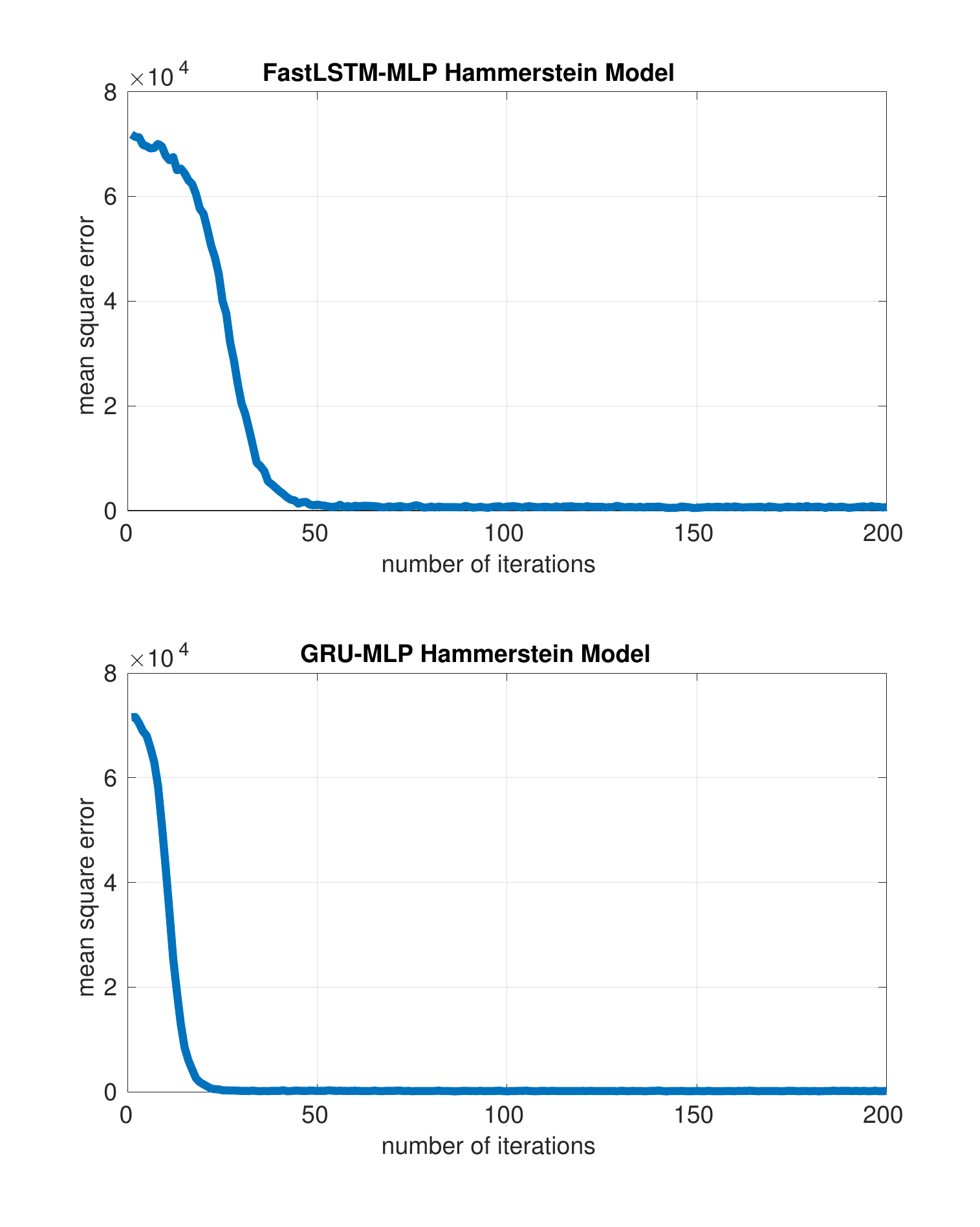}
\caption{\small SISO soft-robot training with FastLSTM/GRU Hammerstein model.}
\label{fig:fastlstm-gru}
\end{figure}

To minimize the complexity of the model structure whilst preserving the effectiveness of the model, we remove the peephole connections of \autoref{fig:lstm} and carry out the same procedure as in $\S$ \ref{sec:lstm-training} with the soft-robot and glassfurnace network. We achieve approximately the same level of convergence (\autoref{fig:fastlstm-gru}) using less parameters in less time (see tables \ref{table:sr-models} \& \ref{table:glassfurnace}) . 
\subsection{Gated Recurrent Units Structure}
The final model structure is the gated recurrent architecture described in\ref{sec:gru}. Like the LSTM models, the structure consists of three nonlinear GRU elements, each followed by $0.35$ drop-out probabilities (to prevent the co-adaptations in training data); this has been shown to lead to better generalization of the NN models \cite{Hinton2014}. The last layer of the GRU structure is a linear dynamic layer that maps the nonlinear states of the system to the head pitch motion. The training algorithm is,
\begin{align}
    \text{Layer 1: GRU }	& 1 \rightarrow 1, 	\, \text{ \{0.35  dropout\}}		\nonumber \\
    \text{Layer 2: GRU }	& 1 \rightarrow 10, \, \text{ \{0.35  dropout\}}		\nonumber \\
    \text{Layer 3: GRU }	& 10 \rightarrow 100, \, \text{ \{0.35  dropout\}}		\nonumber \\
    \text{Layer 4: Linear } & 100 \rightarrow 1
\end{align}
\begin{figure}[tb!]
\centering
\includegraphics[width=3.4in, height=2.5in, keepaspectratio=true]{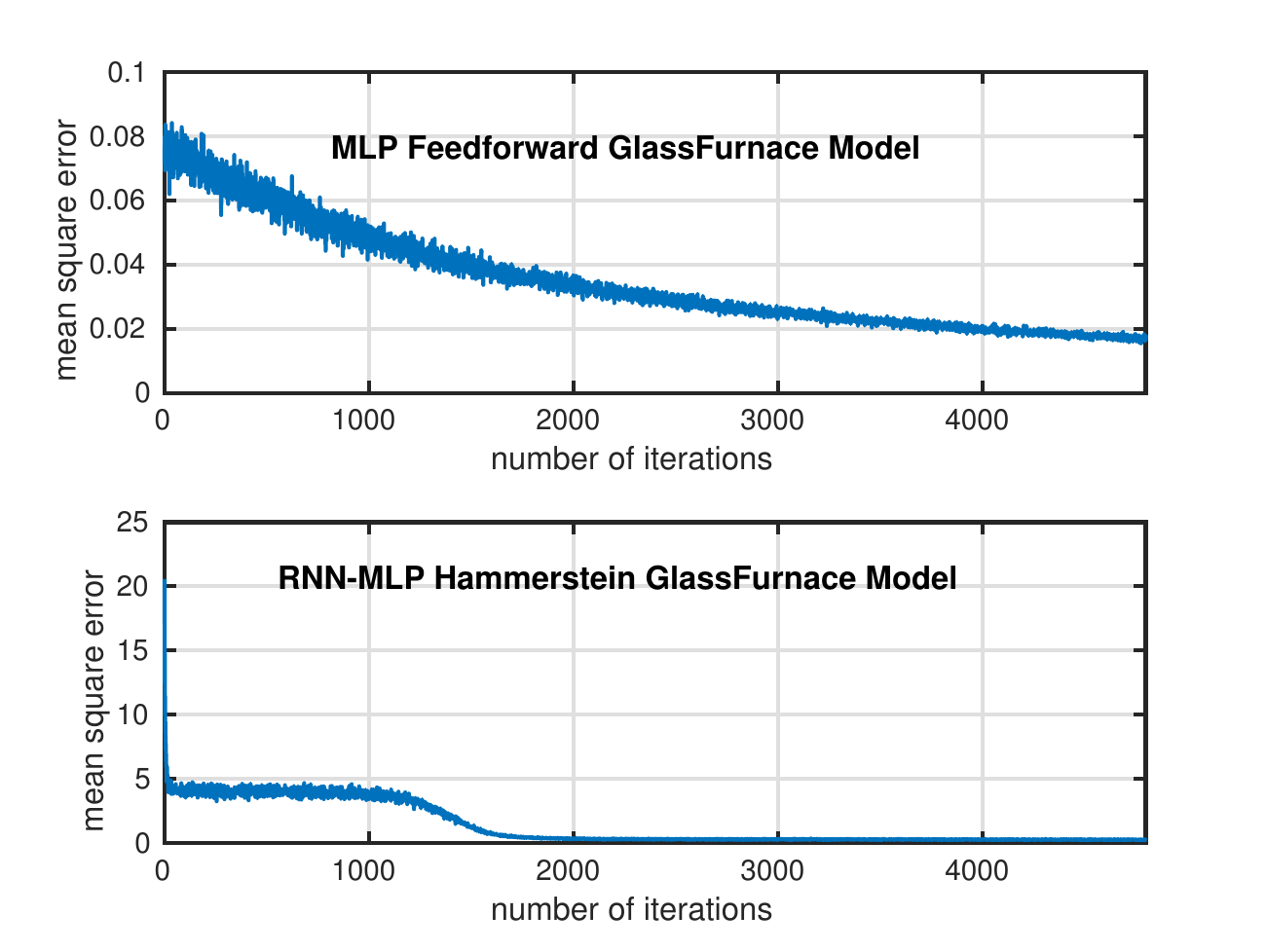}
\caption{\small RNN/Multilayer FeedForward Model Performance on Glassfurnace.}
\label{fig:glass-rnn}
\end{figure}
%
\section{Results and Analyses}
In \autoref{table:sr-models} \& \ref{table:glassfurnace}, the models are characterized by properties that suggest good fit to training data and mean-square losses that are generally acceptable on the given noisy dataset. It is noteworthy that we do not pre-process these datasets nor carry out batch normalization of layers of the network during training. While the multilayer network fits the two datasets well and takes very little time to train, it should be noted that their ability to approximate sequential data may not be robust to model uncertainties and stochastic disturbances as correlated inputs, and self-feedback of input or output information in the dataset are not taken into account by nature of its structure. 

\begin{savenotes}
	\begin{table}[tbph]
		\centering
		\resizebox{\columnwidth}{!}{%
		\begin{tabular}{|c|c|c|c|c|}
			\hline \rule[-2ex]{0pt}{5.5ex} Model   & Estimation Fit (\%)   & Training Time &  $MSE$\footnote{Mean Squared Error} 
			\\ 
			\hline \rule[-2ex]{0pt}{5.5ex}  \textit{MLP} & 99.8471 &  422s &  113.1052 
			\\ 
			\hline \rule[-2ex]{0pt}{5.5ex}  \textit{RNN} & 99.8448 & 1191s  &  101.3624 
			  \\ 
			\hline \rule[-2ex]{0pt}{5.5ex}  \textit{LSTM} & 99.5144 & 876s  &  382.6935 
			\\ 
			\hline \rule[-2ex]{0pt}{5.5ex}  \textit{FastLSTM} & 99.4795 & 881s  & 374.5386  
			\\ 
			\hline \rule[-2ex]{0pt}{5.5ex}  \textit{GRU} & 99.835 & 884s  & 117.0274  
			\\ 
			\hline  
		\end{tabular}
		}
		\caption{Soft Robot Model Performance}
		\label{table:sr-models}
	\end{table}
\end{savenotes}

The charts of Figs. \ref{fig:rnn-lstm} - \ref{fig:glass-rnn} show how well the proposed models capture the dynamics of training data for complicated input-output Borel sets. With proper choice of model hyperparameters and random initialization of weights, each model gradually tunes its parameters to model the dynamics of the underlying system. Deep recurrent neural networks are indeed powerful models that can model almost any input-output relationship as the figures show. 

The fit to estimation data was calculated from 
\begin{align}
\text{Fit}\, ( \%) = \left(1 - \dfrac{\|y - \hat{y} \|}{\|y - \bar{y}\|}\right) \times 100
\end{align}
where $\bar{y}$ is the channel-wise mean and $\|\cdot\|$ is the $2-$norm operator. The mean-square error was calculated according to \eqref{eq:MSE}. 
%
\begin{savenotes}
	\begin{table}[tbph]
		\centering
		\resizebox{\columnwidth}{!}{%
		\begin{tabular}{|c|c|c|c|c|}
			\hline \rule[-2ex]{0pt}{5.5ex} Model   & Estimation Fit (\%)   & Training Time &  $MSE$ 
			\\ 
			\hline \rule[-2ex]{0pt}{5.5ex}  \textit{MLP} & 99.9905 &  462.64s      & 0.015433   
			   \\ 
			\hline \rule[-2ex]{0pt}{5.5ex}  \textit{RNN} & 98.8052 & 726s       &  0.245833    
			   \\ 
			\hline \rule[-2ex]{0pt}{5.5ex}  \textit{LSTM} & 57.1 &  4163.18s      &  3.216   
			   \\ 
			\hline \rule[-2ex]{0pt}{5.5ex}  \textit{FastLSTM} & 91.2822 &   3310.56s     &  0.689027   
			\\ 
			%
			\hline  
		\end{tabular}
		}
		\caption{GlassFurnace (from DaIsY) Model Performance}
		\label{table:glassfurnace}
	\end{table}
\end{savenotes}

From \autoref{table:sr-models} and \ref{table:glassfurnace}, the multi-layer network is the fastest to train giving a decent mean-square error on the dataset in as little as $422$ seconds but as \autoref{fig:mlp-sr} shows, it's approximation capacity is limited by its forward connections-only architecture. While the RNN-MLP Hammerstein models may show a somewhat sluggish performance, the trade-off between training time and robustness of the resulting model could inform the decision of  using the Hammerstein network models against fast-convergence models with little sensitivity to delay, stochastic variables and disturbances. We hypothesize that the Hammerstein models  would perform better in capturing the associated delay from input to output, remembering long-range mappings from input-to-output thus being more robust to tests and deployments. We will investigate the performance of the Hammerstein models on biomedical and mechanical systems in a future work and verify the validity of our assumptions about these models. 

\section{Conclusions}
This work has shown the adaptability of supervised deep network architectures for the identification of nonlinear dynamical systems that are otherwise complicated to model using hand-coded features. Deep networks are easy to train compared to the expert knowledge required in identifying nonlinear regressive models, and they scale well in modeling complicated relationship between input-output data. We designed strictly feedforward and nonlinear Hammerstein model structures for identifying the dynamic relationship between input-output datasets: one gathered from a soft-robot actuator for a motion-alignment correction system in clinical cancer radiotherapy and the other tested on a multi-input and multi-output dataset from DaISy. Through proper hyper-parameters selection, model choice and weights tuning that is appropriate for the learning tasks presented, we demonstrate that complex hand-coding of features characteristic of classical identification can be discarded with deep network-based models. 

With the availability of unit-tested deep network frameworks such as Torch, Tensorflow and Theano, researchers can train datasets with DNNs and generate models that are robust to modeling uncertainties despite complicated structure in data. In future investigations, we will use these models in the real-time identification and control of our proposed soft-robot motion alignment correction systems for H\&N cancer radiotherapy treatments as well as other complex nonlinear phenomena that we are concurrently working with.
%
%
\providecommand\BIBentryALTinterwordstretchfactor{2.5}
\bibliographystyle{IEEEtran}
\bibliography{ref.bib}	 	
\end{document}